\documentclass[11pt]{article}
\usepackage[utf8]{inputenc}
\usepackage{amsmath, amssymb}
\usepackage{graphicx}
\usepackage{caption}
\usepackage{hyperref}
\usepackage{geometry}
\usepackage{enumitem}
\usepackage{titlesec}
\usepackage{authblk,csquotes}
\hypersetup{colorlinks=true,allcolors=blue}

\usepackage{url}
\providecommand{\bibcommenthead}[1]{}
\providecommand{\burl}[1]{\href{#1}{\nolinkurl{#1}}}
\renewcommand{\url}[1]{\href{#1}{\nolinkurl{#1}}}
\providecommand{\urlprefix}{}
\providecommand{\doi}[1]{\href{https://doi.org/#1}{doi:\nolinkurl{#1}}}
\providecommand{\eprint}[2][]{\href{https://arxiv.org/abs/#2}{#2}}
\providecommand{\bibinfo}[2]{#2}
\usepackage{setspace}
\setlist[itemize]{leftmargin=*, itemsep=0.5ex, topsep=0.5ex, parsep=0pt, partopsep=0pt}

\titleformat{\section}{\normalfont\Large\bfseries}{\thesection}{0.5em}{}
\titleformat{\subsection}{\normalfont\large\bfseries}{\thesubsection}{0.5em}{}
\titleformat{\subsubsection}{\normalfont\normalsize\bfseries}{\thesubsubsection}{0.5em}{}

\usepackage{xcolor}

\newcommand{\blue}[1]{\textcolor{blue}{#1}}

\usepackage{soul}
\soulregister\cite7
\soulregister\ref7
\soulregister\onlinecite7

\usepackage{etoolbox}
\renewenvironment{abstract}{%
    \par\vspace{0.5em}%
    \begin{center}\textbf{\Large Abstract}\end{center}\vspace{0.1em}%
    \noindent
}{\par\vspace{0.5em}}

\usepackage{fancyhdr}
\title{\textbf{Dancing Tokens on the Vocabulary Network from \\LLM Generation}}
\author[1,2]{\textbf{Yuan-Chao Hu}\thanks{yuanchao.hu@dimst.ac.cn}}
\affil[1]{Dongguan Institute of Materials Science and Technology, Dongguan, China}
\affil[2]{Songshan Lake Materials Laboratory, Dongguan, China}
\date{\vspace{-1.5em}(Dated: \today)}

\begin{document}

\maketitle
\vspace{-2.5em}

\begin{abstract}
Large language models (LLMs) generate text one token at a time, yet how a particular generation relates to the geometry learned in pre-training remains unclear. The fundamental explainability of LLMs is elusive.
Here, we build a similarity network from the input embeddings of six open LLMs and map generation onto it. The token-embedding network is scale-free while heavy-tailed, as are random embeddings with matched covariance, but far more fragmented. 
Distillation barely changes this topology because input embeddings are largely inherited. The knowledge space is thus deterministic from pre-training.
Consecutive generated tokens are rarely graph neighbours and, in most models, no more often than in shuffled order, so generation is not a walk on the predefined network, although the tokens it visits are concentrated within it. Through the logit lens in a layer-wise manner, tokens settle late in depth in most models, and repetitive models remain confined to small domain-specific sub-graphs.
These observables could serve as inference-time monitors of degeneration. We thus provide a fresh perspective on understanding LLM generation in the microscopic way from node or link prediction by graph learning.
\end{abstract}

\newpage
\begin{spacing}{1.15}

Large language models (LLMs), built on the Transformer architecture~\cite{NIPS2017_attention}, generate text in the autoregressive manner from a prompt. They are now used across science, technology and society~\cite{lecun_deep_2015,AI_science_2023}. Trained on internet-scale corpora and refined by distillation, fine-tuning, reinforcement learning and prompt engineering (chain-of-thought prompting etc.), contemporary models generate fluent and often satisfied text~\cite{deepseek_nature2025,CoT_NIPS2022}. Yet, understanding of how this text is produced has not kept pace with deployment. Scaling and ``densing'' laws describe how capability grows with parameters and data~\cite{xiao_densing_2025}, but they treat the model as a black box. The cost is practical: hallucination~\cite{ji_hallucination_2023,huang_hallucination_2025} cannot be eliminated entirely~\cite{xu_hallucination_2024,kalai_hallucination_2026}, and existing responses --- semantic-entropy uncertainty~\cite{farquhar_semanticentropy_2024}, retrieval-augmented decoding~\cite{lewis_rag_2020} and contrastive-layer decoding~\cite{chuang_dola_2024} --- act on the output without reference to the space in which the model's tokens live.
Acquiring explainability of LLMs is strongly desired but still far from wanted.

A complementary tradition represents knowledge as a network and analyses its topology, tracing back to ancient times. 
Physically, many natural and engineered systems have been found to share scale-free degree distributions and small-world organisation~\cite{barabasi_scaling_1999,watts_smallworld_1998,newman_structure_2003}.
In materials science, networks built from compositions or text have exposed design rules for amorphous alloys~\cite{zhang_constructing_2025,ouyang_graph_2025}. 
Within the field of natural language, a word co-occurrence graph with a scale-free degree distribution was found similarly~\cite{ferrericancho_smallworld_2001}, and word-association networks carry the same statistics~\cite{steyvers_semantic_2005}. Small-world lexico-semantic structure has recently been reported in the input embeddings of LLMs~\cite{liu_wordembeddings_2025}, Lee \textit{et al.} showed that the token embeddings of different LLMs share similar relative orientations~\cite{lee_sharedgeometry_2025}, and hub tokens that organise their neighbours have been traced across layers~\cite{liu_magnets_2026}. Nearest-neighbour graphs in high dimensions, however, develop hubs generically~\cite{radovanovic_hubs_2010}, so heavy-tailed degree statistics require a reasonable comparison before they can be attributed to learning.

In general, LLMs are pre-trained on large corpus by breaking words into tokens at a scale of roughly $10^4$. Combinations of the fragmented tokens will make up texts during inference. The set of tokens per LLM is thus defined before training, while after training each token will be assigned a unique high-dimensional tensorial representation, by optimizing the loss function. These numerical embeddings define the meanings of tokens for computer processing. Finally, three main components function a LLM: the set of tokens, the model architecture, and the parameters (token embeddings and model weights). Given a prompt, the information will be initialized and processed by them to make token-by-token predictions. Nevertheless, the physical mechanism of such a processing is obscure.
This applies to vision-based or multi-modal LLMs in the similar way, sourcing from how tokens are defined.

During LLM generation, there are usually tens of hidden layers, consisting of the attention block and the feed-forward one. After these layer-wise processing, the logit lens will evaluate the possibility of each token as the next chosen one. In principle, these lens are extendable to other hidden layers.
In calculation, it reads out a provisional next-token distribution from each hidden layer~\cite{nostalgebraist_logitlens_2020,belrose_tunedlens_2023}, and the layer beyond which this read-out stops changing has itself become an observable: Lioubashevski \textit{et al.} showed that the top-ranked tokens become fixed in order of their rank~\cite{lioubashevski_top1_2025}, Gupta \textit{et al.} described a guess-then-refine use of depth~\cite{gupta_depth_2025}, and Chen \textit{et al.} related late-layer revision to reasoning effort and accuracy~\cite{chen_deepthinking_2026}. 
The settling layer varies for different tokens and models, which is critical in inference performance.
These read-outs are made token by token, without a global map of the token space on which a whole generation can be located. We therefore ask whether generation can be located on a fixed network built from the pretrained token embeddings, and whether that network constrains the order in which tokens are emitted. 
For LLMs the link between the static geometry of token embeddings~\cite{mikolov_word2vec_2013,ethayarajh_geometry_2019} and the dynamic process of generation remains unestablished. 
Hence, whether a carrier for the complex LLM can be defined as a medium to crack related puzzles from a viewpoint of statistical physics.

In principle, a transformer-based language model learns the embeddings of a set of tokens from its training corpora by capturing the hidden correlations among them through the self-attention mechanism. Given user input, the model transforms the sequential embeddings into a probability distribution over the token set to predict the next token~\cite{NIPS2017_attention}. 
The temperature will determine how stochastic this selection is.
This is the inference process from the pre-trained model. The intermediate computations are high-dimensional and entangled. The mapping from input context to output tokens resists the kind of mechanistic account that is routine in the natural sciences, for example, in the field of complex disordered systems. The data flow within the architecture during a specific generation has received comparatively little study. Token prediction is intrinsically a search over the vocabulary, driven by the conditional probabilities that the architecture computes: transformer layers capture the hidden correlations between tokens, while multi-layer perceptrons compute those between the features of tokens. An efficient search requires a well-defined representation of these tokens and their features, and understanding how the search proceeds is a prerequisite for principled domain adaptation.

In this study, we construct, for each of six open LLMs of 7--21 billion parameters from four developers, a token-similarity network from the pretrained input embeddings, fix its cutoff at the connectivity reference point where the mean degree crosses unity, and map autoregressive generation onto it. Even though the cutoff will quantitatively change the network properties, here our main purpose is to propose the viewpoint of understanding LLM generation from the token graph. How the cutoff will lead these quantitative variations is an open interesting question for future investigation.
Currently, we find that the networks are heavy-tailed but far more fragmented than random-embedding nulls, even after surface-form variants are merged. Their topology fingerprints the model family (i.e. model architecture) and survives distillation, because the input embeddings are largely inherited. This raise the question what distillation really changes to modify the performance of LLMs in certain domains. Consecutive generated tokens are not network neighbours, although a generation occupies a concentrated region of the network. Tokens settle late in depth in most models. The breadth of coverage separates repetitive from exploratory generation. These findings pave a way to explaining LLM generation by graph learning on a quantitative token network.

\section*{Results}

\subsection*{Network representation of LLM generation}

\begin{figure}[t!]
\centering
\includegraphics[width=\textwidth]{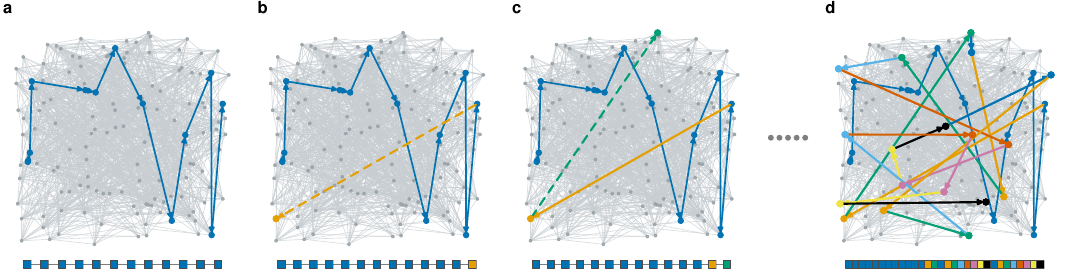}
\caption{\textbf{Schematic of autoregressive generation mapped onto a token graph.}
\textbf{a}, The tokens of an input sequence are located as nodes of the token-similarity network (grey edges), and their order is drawn as directed sequence steps (blue arrows); sequence steps need not coincide with similarity edges.
\textbf{b}, The predicted next token is located as a further node, joined to the sequence by a new step (orange; dashed, predicted step).
\textbf{c}, The extended sequence becomes the new input (orange step now part of the sequence) and the following token is predicted (green, dashed).
\textbf{d}, Iterating \textbf{a}$\rightarrow$\textbf{c} maps the full generation onto the network (one colour per generated step). Linked squares beneath each panel show the corresponding input (blue) and generated (coloured) token sequence.}
\label{fig1}
\end{figure}

We represent a model in its token space as a network: every token in the vocabulary is a node, and two nodes are connected when the cosine similarity of their input embeddings exceeds a threshold $\mathcal{S}$ (Methods). A generation is then mapped onto this fixed network by locating the prompt and generated tokens as nodes and drawing the order of emission as directed steps between them (Fig.~\ref{fig1}). Nothing in the construction requires a sequence step to be a similarity edge, which is a static property of a LLM as a pre-training by-product. The network thus encodes the static geometry laid down in pre-training, and the sequence the choices made at inference.
The token graph is fixed once pre-training ends, whereas the generated sequence is produced at inference, so the construction separates a quenched object (the network) from an annealed one (the sequence). 

\begin{figure}[t!]
\centering
\includegraphics[width=\textwidth]{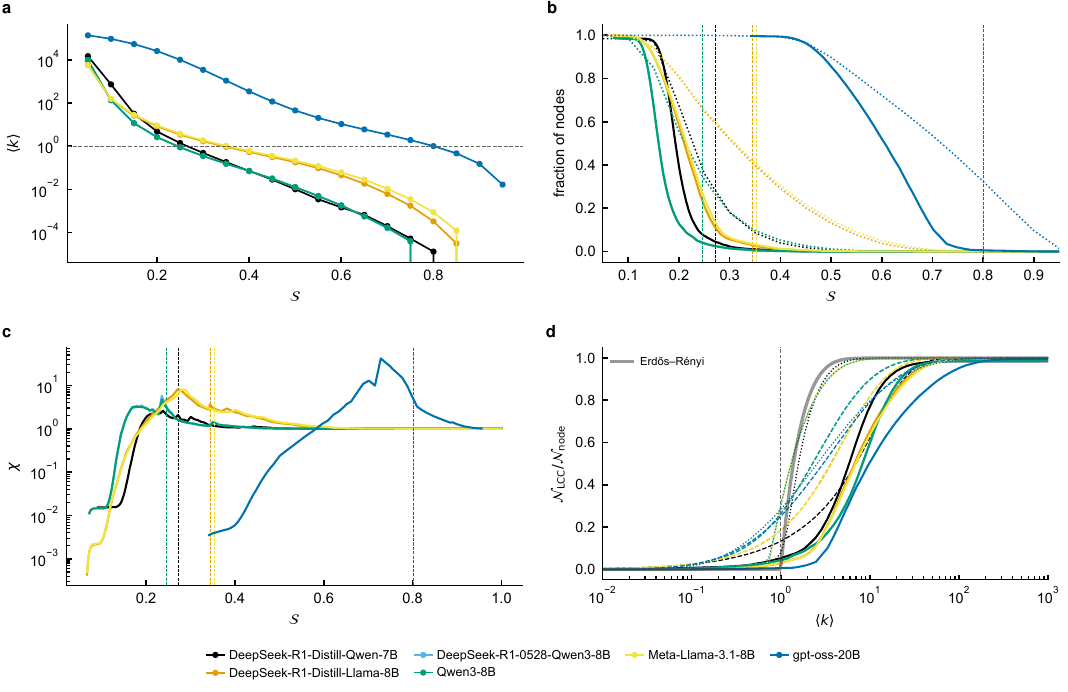}
\caption{\textbf{Network construction, the reference cutoff and percolation.}
Each curve is one model (legend below); the cosine-similarity threshold $\mathcal{S}$ is swept from 0.05 to 0.95.
\textbf{a}, Mean degree $\langle k\rangle$ versus $\mathcal{S}$ (log scale); the dashed line marks $\langle k\rangle=1$, equivalent to $\mathcal{N}_\mathrm{edge}=\mathcal{N}_\mathrm{node}/2$ with edges counted as unordered pairs.
\textbf{b}, Fraction of nodes in the largest connected component (solid) and fraction of nodes with degree $>0$ (dotted) versus $\mathcal{S}$; vertical dashed lines mark each model's $\mathcal{S}_\mathrm{c}$, where $\langle k\rangle=1$.
\textbf{c}, Susceptibility $\chi$, the mean size of all clusters other than the largest, versus $\mathcal{S}$ (log scale), with $\mathcal{S}_\mathrm{c}$ marked as in \textbf{b}.
\textbf{d}, Largest-component fraction versus $\langle k\rangle$ for each model (solid), its Gaussian null (dashed) and its shuffled-dimension null (dotted), with the Erd\H{o}s--R\'enyi prediction (thick grey) and $\langle k\rangle=1$ (vertical dashed line).}
\label{fig2}
\end{figure}

We study the six models of Table~\ref{tab:models}, with vocabularies of about 128--201k tokens: one pretrained base model (Meta-Llama-3.1-8B), three distilled models from DeepSeek-R1 reasoning traces and two post-trained instruction or reasoning models. 
These models are open-sourced and we hope to choose them to cover different model architectures (\blue{Supplementary Note~1}).
The threshold $\mathcal{S}$ is the single free parameter of the construction, and no value is correct a priori (\blue{Supplementary Fig.~1}), so we swept it from 0.05 to 0.95. The mean degree $\langle k\rangle$ passes through unity within a narrow window of $\mathcal{S}$ for all models (Fig.~\ref{fig2}a), and the largest connected component (LCC) grows as the cutoff is lowered through this window (Fig.~\ref{fig2}b). The curves group by model family: gpt-oss-20B has the highest token similarities and remains essentially fully connected (LCC above $99\%$) up to $\mathcal{S}\sim0.4$, whereas the Llama and Qwen families collapse onto two groups with much smaller similarities (\blue{Supplementary Fig.~1}).

\begin{table}[b!]
\centering
\small
\caption{The six studied LLMs and their key architectural parameters. $\mathcal{N}_\mathrm{tot}$ is the total parameter count; $L$ the number of transformer blocks (layers); $d$ the hidden size; heads/KV the number of attention and key--value heads; $\mathcal{V}$ the vocabulary size. All use SiLU activations and root-mean-square layer normalisation (RMSNorm). gpt-oss-20B is a mixture-of-experts (MoE) model with 32 experts (4 active per token). 
}
\resizebox{\textwidth}{!}{%
\begin{tabular}{l l l r r r r r}
\hline
Model & Type (starting checkpoint) & Architecture & $\mathcal{N}_\mathrm{tot}$ & $L$ & $d$ & heads/KV & $\mathcal{V}$ \\
\hline
Meta-Llama-3.1-8B            & pretrained base & Llama           & 8.0\,B  & 32 & 4096 & 32/8 & 128256 \\
Qwen3-8B                     & post-trained, hybrid thinking (Qwen3-8B-Base) & Qwen3           & 8.2\,B  & 36 & 4096 & 32/8 & 151936 \\
DeepSeek-R1-0528-Qwen3-8B    & R1-0528 distilled (Qwen3-8B-Base) & Qwen3           & 8.2\,B  & 36 & 4096 & 32/8 & 151936 \\
DeepSeek-R1-Distill-Llama-8B & R1 distilled (Llama-3.1-8B) & Llama           & 8.0\,B  & 32 & 4096 & 32/8 & 128256 \\
DeepSeek-R1-Distill-Qwen-7B  & R1 distilled (Qwen2.5-Math-7B) & Qwen2           & 7.6\,B  & 28 & 3584 & 28/4 & 152064 \\
gpt-oss-20B                  & reasoning, post-trained & GPT-OSS (MoE)   & 20.9\,B & 24 & 2880 & 64/8 & 201088 \\
\hline
\end{tabular}}
\label{tab:models}
\end{table}

We use the onset of connectivity to fix the cutoff. In an Erd\H{o}s--R\'enyi random graph a giant component emerges at $\langle k\rangle = 1$, when the number of undirected edges reaches half the number of nodes~\cite{newman_structure_2003,erdos_renyi_1960}, and we denote by $\mathcal{S}_\mathrm{c}$ the cutoff at which our networks satisty this condition. 
We borrow this cutoff for the finite systems here, which is well defined, reproducible and model specific.
The token graphs do not, however, percolate like a random graph. This renders possible physical distinctions of the designed graph from the random one, hypothesizing the mysterious messages under the construction.  At $\mathcal{S}_\mathrm{c}$ the LCC holds only $0.6$--$5\%$ of the vocabulary, and it reaches half of the vocabulary only at $\langle k\rangle\approx6$--$11$, against $1.4$ for an Erd\H{o}s--R\'enyi graph; the susceptibility, the mean size of the finite clusters, peaks at $\langle k\rangle=1.2$--$2.5$ (Fig.~\ref{fig2}b--d). $\mathcal{S}_\mathrm{c}$ is therefore a reproducible, model-specific reference point near the onset of connectivity rather than a sharp transition, and we perform all subsequent analysis at each model's working cutoff $\mathcal{S}^{*}$, a sweep grid point $0.015$--$0.052$ below $\mathcal{S}_\mathrm{c}$ (Methods). This criterion differs from the minimum-connectivity rule of Liu \textit{et al.}~\cite{liu_wordembeddings_2025} and the training-time percolation of Lubana \textit{et al.}~\cite{lubana_percolation_2024} (\blue{Supplementary Note~3}).

Three groups of behaviour are visible in Fig.~\ref{fig2}, and they follow the model architecture. gpt-oss-20B (MoE) has the highest token similarities and the most persistent connectivity: it has the largest $\langle k\rangle$ (and hence $\mathcal{N}_\mathrm{edge}$) among all the models, and its fully connected graph survives up to $\mathcal{S}\sim0.4$. The curves collapse within another two groups: the Llama architecture and the Qwen architecture. Their token similarities are much smaller than those of gpt-oss-20B, as shown by the fast decay of $\langle k\rangle$ and of the LCC in Fig.~\ref{fig2}a,b. What distillation does change remains a question for future work, while it barely influences the token graph from pretraining (see, for example, ref.~\cite{cloud_subliminal_2026}). This observation rationalizes the necessity of pretraining LLM for different domains.

\subsection*{Physical properties of token graphs}

\begin{figure}[t!]
\centering
\includegraphics[width=\textwidth]{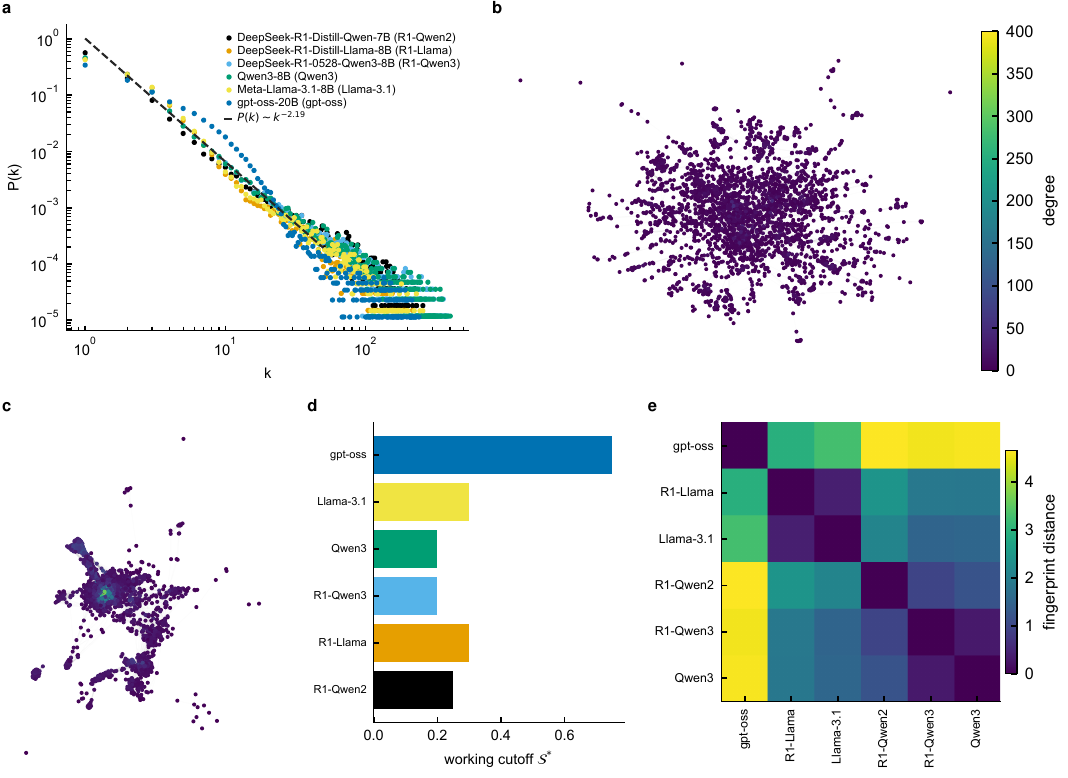}
\caption{\textbf{Network fingerprints across models.}
\textbf{a}, Degree distribution $P(k)$ at each model's working cutoff $\mathcal{S}^{*}$ (log--log; see \textbf{d}), restricted to nodes with $k\ge1$ and normalised over them. Abbreviated model names in the legend are used in panels \textbf{d} and \textbf{e}.
\textbf{b,c}, Largest connected component coloured by node degree for gpt-oss-20B (\textbf{b}) and the top-degree core of Qwen3-8B (its 3,000 highest-degree nodes; \textbf{c}), on a shared degree scale (colour bar in \textbf{b}).
\textbf{d}, Working cutoff $\mathcal{S}^{*}$ per model.
\textbf{e}, Pairwise Euclidean distance between model network fingerprints (mean degree, modularity and community fraction of the LCC)}
\label{fig3}
\end{figure}

At the working cutoff the degree distribution is heavy-tailed for every model (Fig.~\ref{fig3}a), giving $P(k)\sim k^{-\gamma}$ with $\gamma=2.19\pm0.04$.\blue{(Supplementary Table~3 and Supplementary Fig.~2)}. 
The tail exponent near what from a word co-occurrence graph, approaching the Barab\'asi--Albert value expected under preferential attachment~\cite{barabasi_scaling_1999,ferrericancho_smallworld_2001}. These successes motivate an analogous treatment of language models, in which the token embeddings learned during pre-training are examined for network structure and for its role in generation.
Suprisingly, a Gaussian with each model's mean and covariance, thresholded at the same $\langle k\rangle$, reproduces a comparable tail (\blue{Supplementary Fig.~2}), as expected from the hubness of high-dimensional data~\cite{radovanovic_hubs_2010}. What separates the trained embeddings from their nulls is fragmentation: at matched $\langle k\rangle$ the LCC holds $2$--$13\%$ of the tokens for the models, against $17$--$55\%$ for the Gaussian null and $36$--$83\%$ for a null with each embedding dimension permuted independently, while the fraction of tokens with at least one neighbour is similar (\blue{Supplementary Note~6}). The linked tokens of a trained model sit in many small clusters rather than in one giant component (Fig.~\ref{fig2}d). This spatial distribution features the token graph, indicating strongly heterogeneous dynamics during model training and inference.
These observations define the uniqueness of the token graphs from LLM pretraining, which may serve as a typical system for network study.

Decoding the vocabulary shows what these hubs and clusters are. In all six models, at least 99 of the 100 highest-degree tokens contain no letter or digit. They are closing brackets, semicolons and quotation marks fused with newline or carriage-return characters (\blue{Supplementary Table~5}). Surface-form duplication also shapes the local structure: $26$--$42\%$ of the edges join tokens that differ only in case or surrounding whitespace, and $34$--$45\%$ of the finite clusters consist solely of such variants. The fragmentation nevertheless survives their removal. Restricted to tokens that contain a letter, the LCC holds $1.9$--$8.7\%$ of those tokens, against $11$--$35\%$ for a Gaussian null fitted to them at the same edge count. With all surface-form variants of a string merged into one node, it holds $2.9$--$11.9\%$ of the merged nodes, against $22$--$63\%$ for the Gaussian null merged in the same way (\blue{Supplementary Note~6}). Fragmentation is thus a property of the trained tokenizer--embedding pair, part of which reflects the byte-pair encoding of one string in several forms. What the remaining clusters encode is not resolved here.

The network topology also fingerprints the model. We summarised each network by the mean degree, modularity and community fraction of its largest connected component and computed the pairwise distance between models (Fig.~\ref{fig3}b--e; Methods). Models of the same family cluster together and gpt-oss-20B is the clear outlier. The fingerprint distance between DeepSeek-R1-0528-Qwen3-8B and Qwen3-8B, two siblings post-trained from Qwen3-8B-Base, is nearly seven-fold smaller than the mean distance from DeepSeek-R1-0528-Qwen3-8B to the four other models. This invariance is inherited through the weights. The distilled checkpoints keep their family's tokenizer and largely its input embeddings: the relative Frobenius change of the embedding matrix is $0.11$ between the two Qwen3-8B siblings and $0.14$ between DeepSeek-R1-Distill-Llama-8B and its base, Meta-Llama-3.1-8B, with mean row-wise cosines of $0.995$ and $0.988$ (Supplementary Note~6). The fingerprint is therefore set by the model family and its pre-training, which distillation and reinforcement learning barely move.
We note that the LCC structure differs visibly between gpt-oss-20B and Qwen3-8B (Fig.~\ref{fig3}b,c). Growth by preferential attachment to emerging hubs is one hypothesis, but the Gaussian null could reproduce a comparable tail without any growth process, which makes its origin elusive.
\subsection*{Generation order relative to the network}

\begin{figure}[t!]
\centering
\includegraphics[width=\textwidth]{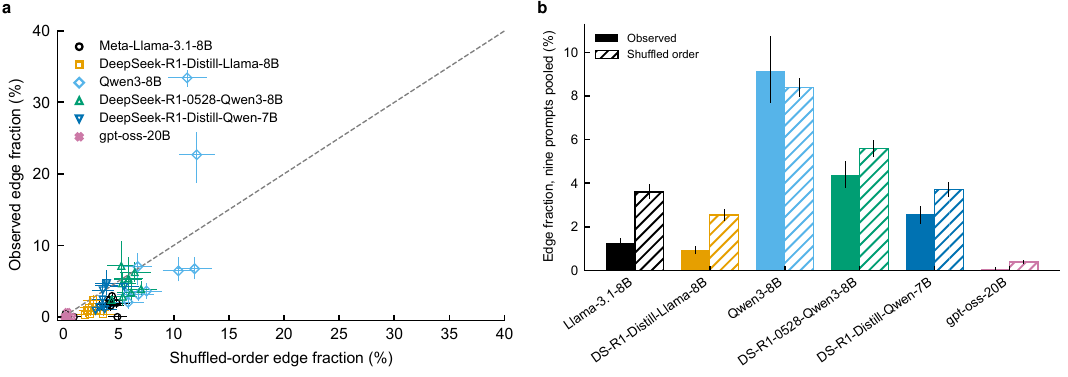}
\caption{\textbf{Transition-edge test for consecutive generated tokens.}
\textbf{a}, Percentage of consecutive generated-token pairs that are edges of the token network at $\mathcal{S}^{*}$ (observed) against the same percentage for the tokens of the run in shuffled order, one point per model and prompt (54 greedy-decoded runs; marker and colour identify the model, legend). Vertical bars, 95\% moving-block bootstrap intervals of the observed value; horizontal bars, 2.5--97.5 percentile range over 1,000 shuffles; dashed line, equality.
\textbf{b}, Observed (filled) and shuffled-order (hatched) edge percentage per model.
Immediate repeats of a token are excluded from both.}
\label{fig:transition}
\end{figure}

With the network fixed, we asked whether generation moves along its edges. 
We note from the transformer attention mechanism that essentially the token graph should be dynamical with attention scores changing based on the prompt text for the next token. Nevertheless, to understand the graph, we first take the static graph from pre-training for early mining.
For every pair of consecutive generated tokens we tested whether the two are neighbours at $\mathcal{S}^{*}$, excluding immediate repeats of a token, and compared this edge fraction with that of the same tokens in shuffled order, which keeps the composition of the run and destroys only its order (Methods). Consecutive tokens are rarely neighbours (Fig.~\ref{fig:transition}a). Pooled over 9 prompts (about $10^4$ transitions per model), the edge fraction ranges from $0.07\%$ for gpt-oss-20B to $9.1\%$ for Qwen3-8B, and for five of the six models it lies below the shuffled-order value (Fig.~\ref{fig:transition}b; \blue{Supplementary Table~4}). The exception is Qwen3-8B ($9.1\%$, $95\%$ confidence interval $7.7$--$10.8\%$, against $8.4\%$ shuffled), because two movie-prompt runs that repeat lists of titles reach $23\%$ and $33\%$. Of the 54 runs, 36 fall below the $95\%$ range of their shuffled baseline and three lie above it. Sampling does not change the picture: in five temperature-$0.6$ runs each of Qwen3-8B and gpt-oss-20B, the edge fraction is $5.3$--$7.5\%$ and $0\%$, against shuffled-order values of $9.7$--$11.8\%$ and $0.3$--$0.5\%$ (\blue{Supplementary Note~6}). Generation is therefore not a walk on the static similarity network but a dynamical one: the next token is, as a rule, not an embedding neighbour of the current one. The graph is dynamical as assumed from the attention mechanism. This also raises a question that whether a specific domain will have a unique graph layout. The layout of the generation features will also determine the generation quality, as discussed below.

The network nevertheless locates the generation. The shuffled-order baseline, which measures how often two tokens of the same run are neighbours irrespective of order, exceeds the density of the whole graph by a factor of about $400$ for gpt-oss-20B and $1{,}900$--$4{,}900$ for the other models (\blue{Supplementary Table~4}), so the tokens emitted for one prompt occupy a small, densely linked part of the network, partly because runs reuse the punctuation hubs identified above. This suggests one condition to optimize the pretraining process.
An intervention supports this reading. Rotating every input embedding of Qwen3-8B by the same small angle (row cosine $0.99$ to the original), either towards a random direction or away from its network neighbours, keeps the model fluent; the targeted edit deletes $92\%$ of the network's edges (mean degree $2.62\rightarrow0.20$) and the random edit $9\%$, yet on two prompts the generations they produce are indistinguishable in edge fraction, concentration, repetition and settling depth, all measured on the original network (\blue{Supplementary Note~7; Supplementary Table~6}). We therefore use the network as a coordinate system onto which generation is mapped, not as its generator. The static graph provides a new carrier to tackle this complex puzzle.
The drawn paths of Figs.~\ref{fig4}e,f and~\ref{fig:quality}a,b join consecutive tokens by sequence steps that are, in general, not similarity edges.

\subsection*{Layered trajectories of token generation}

\begin{figure}[t!]
\centering
\includegraphics[width=\textwidth,height=0.7\textheight,keepaspectratio]{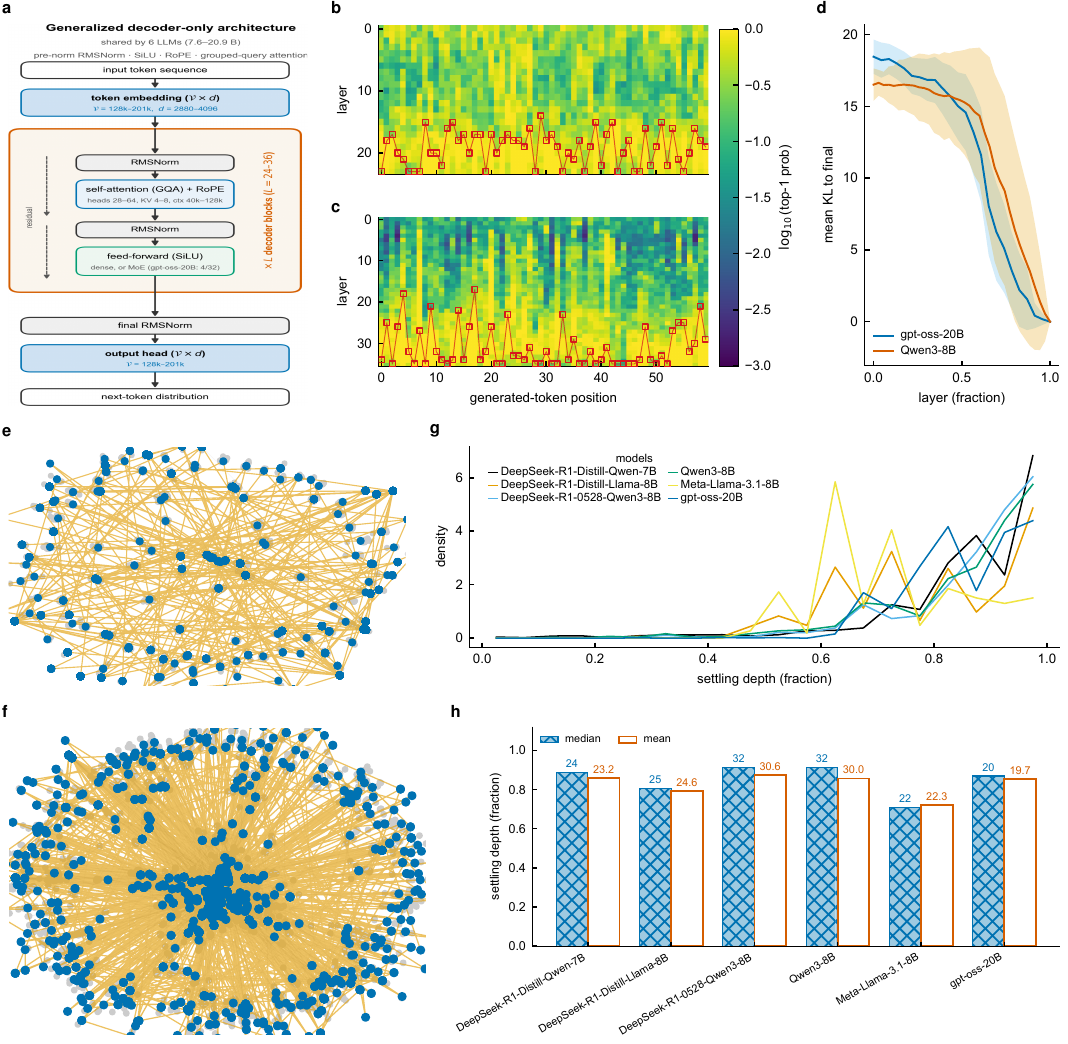}
\caption{\textbf{Model architecture and the layered generation trajectory (gpt-oss-20B and Qwen3-8B).}
\textbf{a}, Schematic generalised decoder-only architecture shared by the six studied models. 
\textbf{b,c}, Logit-lens top-1 probability 
for gpt-oss-20B (\textbf{b}) and Qwen3-8B (\textbf{c}) over the first 60 generated tokens of the simple machine-learning prompt. Red squares linked by a line mark each token's settling layer. 
\textbf{d}, Mean Kullback--Leibler divergence from each layer's distribution to the final-layer distribution versus normalised depth $\ell/(L-1)$ (gpt-oss-20B, blue; Qwen3-8B, vermillion; shaded bands, $\pm1$ s.d. across positions).
\textbf{e,f}, Generated sequence mapped onto the prompt-and-generated token sub-graph for gpt-oss-20B (\textbf{e}) and Qwen3-8B (\textbf{f}): similarity edges in grey, visited tokens in blue and sequence steps between consecutive tokens in orange.
\textbf{g}, Distribution of settling depth (settling layer divided by $L-1$) for all six models (legend in-panel).
\textbf{h}, Median (filled, cross-hatched) and mean (open) settling depth per model, normalised by $L-1$; the absolute median and mean settling layer are printed above each bar.}
\label{fig4}
\end{figure}

We next followed the data flow within the architecture (Fig.~\ref{fig4}a; \blue{Supplementary Note~1}) using the logit lens, projecting the hidden state of every layer through the shared output head (Methods). For each token we define its settling layer as the earliest layer at and beyond which the layer-wise top-1 prediction matches that of the deepest layer and remains stable; under greedy decoding this target is the emitted token. The settling layer is the saturation event of Lioubashevski \textit{et al.}~\cite{lioubashevski_top1_2025}, and it compresses into one number per token the guess-then-refine use of depth described by Gupta \textit{et al.}~\cite{gupta_depth_2025}; Chen \textit{et al.} counted tokens still revised in late layers of GPT-OSS, DeepSeek-R1 and Qwen3 models and found that their share tracks accuracy~\cite{chen_deepthinking_2026}. With the settling layer quantified, we analyze its distribution across a panel containing base, distilled and reasoning-tuned members of the same families, and its ordering relative to the entropy of the lens distribution.

Tokens settle late in most models. The share of tokens settling within the top quartile of layers ranges from $33\%$ (Meta-Llama-3.1-8B) to $85\%$ (gpt-oss-20B) and exceeds one half in five of the six models; for gpt-oss-20B the median settling depth is $0.87$ of the stack, and tokens settle only once the lens prediction has become confident (Fig.~\ref{fig4}b,c,g). For five of the six models the mean settling depth lies below the median (Fig.~\ref{fig4}h), so a thin tail of early-settling tokens coexists with a bulk that resolves near the top of the stack, and the divergence to the final distribution decreases gradually with depth (Fig.~\ref{fig4}d). Mapped onto the network, the sequence runs through a structured region (Fig.~\ref{fig4}e,f; per-model trajectories in \blue{Supplementary Fig.~3}), i.e. a sub-graph of the token graph. In the dense models, the layers above the mean settling layer hold $7$--$24\%$ of the parameters, varying by up to $1.8$-fold across prompt domains. This is an upper bound on the average depth that an ideal per-token early exit~\cite{schuster_calm_2022} could skip to save computational resources, and, consistent with reports that early exit pays least in newer and tuned models~\cite{wei_earlyexit_2026}, the base model carries the largest fraction (\blue{Supplementary Note~4; Supplementary Table~2 and Supplementary Fig.~5}). This may rationalize the advantage of the mixture-of-experts design in LLMs.

Beneath this convergence, the Shannon entropy of the lens distribution first rises and then falls with depth (\blue{Supplementary Fig.~4 and Supplementary Note~8}), reproducing the expansion-then-pruning profile of Ali \textit{et al.}~\cite{ali_entropylens_2025}, and its falling branch collapses onto a common logistic curve after rescaling for gpt-oss-20B and Qwen3-8B (median per-token $R^2\approx0.91$). Notably, this entropy collapse sits mid-stack, ahead of the late settling of the top-1 identity. Combined with the ordered saturation of Lioubashevski \textit{et al.}, in which lower-ranked candidates are fixed after the top-1~\cite{lioubashevski_top1_2025}. This suggests a three-step sequence in depth: the distribution concentrates, the top-1 identity commits and the runners-up follow.


We also compare the network information for three fields by designing the prompts, i.e. metallic-glass, machine-learning and movie (\blue{Supplementary Table 7 and Supplementary Fig.~6}).
The three domain networks differ markedly in their nodes and community structure for both gpt-oss-20B and Qwen3-8B. The cross-domain overlap of activated tokens is low (Jaccard $0.07$--$0.10$ for gpt-oss-20B and $0.08$--$0.13$ for Qwen3-8B). Averaged over the six models, $70\%$ of the tokens generated for a given domain are never used in the other two (metallic glasses $68\%$, machine learning $71\%$, movies $70\%$;). Each domain resolves into its own communities: for gpt-oss-20B, the metallic-glass, machine-learning and movies sub-graphs contain $61$, $49$ and $108$ communities, respectively. 
Because each field activates a largely disjoint set of high-similarity token clusters, its functional sub-graph is field-intrinsic. Pruning a model to that sub-network is left to future work. 
This indicates the feasibility of developing domain-specific LLMs over the general intelligent ones, from the perspective of token graphs.

\subsection*{Repetition as confinement on the network}

\begin{figure}[t!]
\centering
\includegraphics[width=\textwidth]{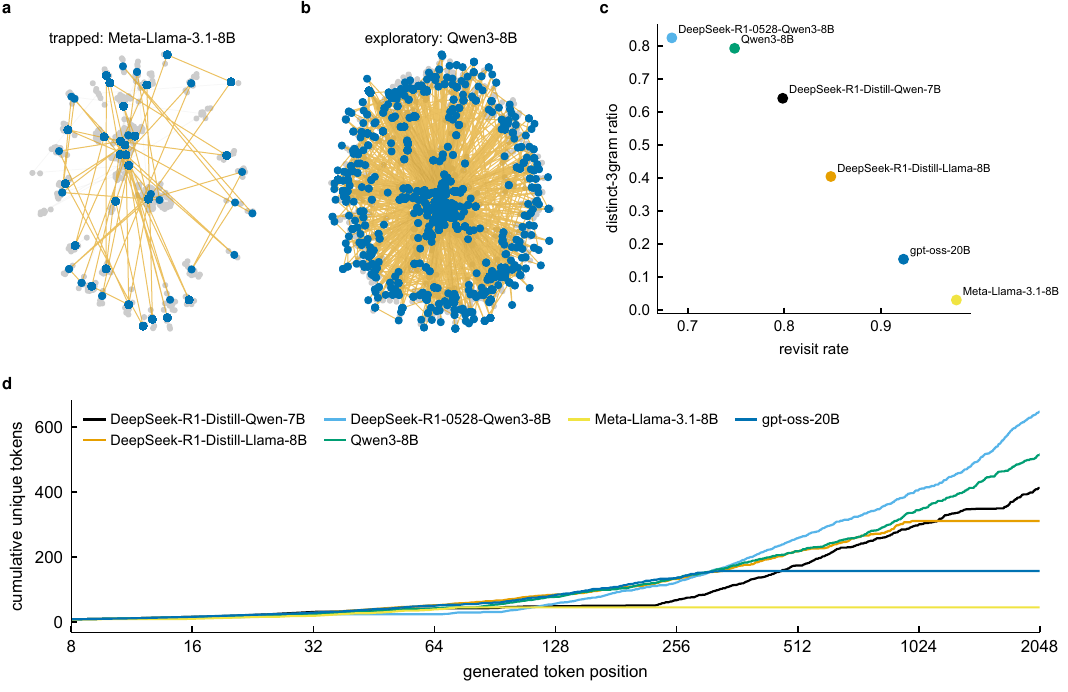}
\caption{\textbf{Repetition and exploration mapped onto the token network.}
\textbf{a,b}, Generated sequence mapped onto the prompt-and-generated token sub-graph (similarity edges in grey, visited tokens in blue, sequence steps in orange) for a repetitive model (Meta-Llama-3.1-8B, \textbf{a}) and an exploratory model (Qwen3-8B, \textbf{b}).
\textbf{c}, Token-revisit rate versus distinct-3-gram ratio for all six models.
\textbf{d}, Cumulative number of unique tokens versus generated-token position over the LLM generation run.} 
\label{fig:quality}
\end{figure}

Finally, we asked how broadly each model's generation covers the network. Meta-Llama-3.1-8B collapses into a tight cluster (Fig.~\ref{fig:quality}a), whereas Qwen3-8B spreads across many communities (Fig.~\ref{fig:quality}b). Across the six models, a higher token-revisit rate accompanies a lower diversity of generated $n$-grams (Pearson $r=-0.98$, $p\approx5\times10^{-4}$; Fig.~\ref{fig:quality}c). Both axes are repetition statistics of the same sequence, so their correlation is expected, and neither measures correctness; the informative content is the ordering of the models, which is stable across prompts ($r$ between $-0.93$ and $-1.00$ over nine prompts; \blue{Supplementary Note~6}).
It measures the generation quality of the LLM regarding to the given prompt. A larger revisit rate indicates a deeper trap over fewer tokens in the high-dimensional landscape of the learning dynamics.

Regarding the model comparison, Meta-Llama-3.1-8B is a pretrained base model, the other five are post-trained. All runs used greedy decoding of raw-text prompts without a chat template (Methods), a regime that the Qwen3 developers advise against because it can cause endless repetition. The base model does not drive the correlation: without it, $r$ lies between $-0.92$ and $-0.99$ over the nine prompts ($n=5$; $p\le0.029$ on every prompt). Nor does post-training guarantee exploration: the post-trained gpt-oss-20B and Qwen3-8B are each the most repetitive model on one of the nine prompts. The ordering is a property of each checkpoint under this regime, not of a model type. This highlights the necessity of efficient prompt engineering or reasoning process to guide model output.

The accumulation of new tokens makes the confinement explicit (Fig.~\ref{fig:quality}d). Over a 2{,}048-token run, Meta-Llama-3.1-8B plateaus at $45$ distinct tokens, having fallen into a short cycle, whereas DeepSeek-R1-0528-Qwen3-8B reaches $648$. The two same-family pairs isolate post-training more cleanly: DeepSeek-R1-Distill-Llama-8B, distilled from Meta-Llama-3.1-8B, reaches $311$ distinct tokens, seven-fold more than its base, and on the Qwen3-8B-Base backbone the reasoning-distilled sibling reaches $648$ against $515$ for Qwen3-8B; in both pairs the distilled model revisits less on all nine prompts. 
Hence, the bare difference made by distillation onto the input embeddings or network topology is actually beneficial to leverage the model performance by dragging it out of the deep minima it has been trapped in.

Sampling moderates the confinement of some models: at temperature $0.6$ the revisit rate of gpt-oss-20B falls from $0.90$ to $0.44$--$0.82$ across five seeds (three of which stop early at an end-of-sequence token), whereas that of Qwen3-8B barely changes ($0.72$--$0.77$ against $0.77$; \blue{Supplementary Note~6}). Early saturation of token discovery is thus a simple readout of confinement to a sub-graph, paralleling the geometric account of mode collapse by Du and Tanaka-Ishii~\cite{du_modecollapse_2026} and the attractor cycles of Wang \textit{et al.}~\cite{wang_attractor_2025}. This is also a new feature to fingerprint a model.

\section*{Discussion}

Our central result is that the pretrained input embeddings of an LLM define a fragmented token network onto which generation can be mapped, and that generation does not move along its edges. The transition-edge test separates two readings that a drawn path conflates. As a generator, the similarity graph would predict the next token from the current one, and it does not (Fig.~\ref{fig:transition}). As a coordinate system, it places the tokens of a generation in a concentrated, domain-specific region and exposes how broadly a model's output covers it (Figs.~\ref{fig:transition} and~\ref{fig:quality}). That consecutive tokens are less often neighbours than random pairs from the same run is consistent with fluent text alternating between function words, punctuation and content words, which the graph places in different clusters. In physical terms, the embedding geometry is a quenched variable, frozen at the end of pre-training, whereas the sequence is annealed at inference: post-training leaves the embedding geometry nearly unchanged (Fig.~\ref{fig3}e) while changing the generated sequences (Fig.~\ref{fig:quality}d).

The static results temper the analogy with language networks. The degree exponent resembles the scale-free class that for word co-occurrence~\cite{ferrericancho_smallworld_2001}. But a matched Gaussian reproduces it, consistent with high-dimensional hubness~\cite{radovanovic_hubs_2010} and with geometry shared across LLM embeddings~\cite{lee_sharedgeometry_2025}. In all six models the hubs are punctuation--newline tokens rather than words. For $2<\gamma<3$ the percolation threshold of a random graph with the measured degree sequence vanishes in the thermodynamic limit~\cite{barabasi_scaling_1999,newman_powerlaw_2005}, so $\mathcal{S}_\mathrm{c}$ is a reference point for a finite system rather than a critical point, where the largest component holds half the vocabulary. 
What the trained embeddings add is fragmentation into many small clusters, which the nulls lack and which survives the removal of surface-form duplicates. The distillation invariance, in turn, is inheritance: the distilled checkpoints differ from their closest relatives by a relative Frobenius change of only $0.11$--$0.14$ in their input embeddings, so the fingerprint reports how little post-training moves the embedding geometry, much as instruction tuning, unlike scale, leaves the alignment of LLMs with human language processing unchanged~\cite{gao_brainalignment_2025}. Unlike provenance detectors that recover lineage from behaviour~\cite{wu_llmdna_2026} or gradients~\cite{wu_tensorguard_2025}, it describes the geometry onto which generation is mapped.

On the dynamic side, from settling layer analsyis~\cite{lioubashevski_top1_2025,gupta_depth_2025,chen_deepthinking_2026}, and we identify several remarkable features: late settling in most models, least so in the base model, and entropy concentration ahead of identity commitment. The breadth of coverage, by contrast, is a network observable. Spatial collapse, a high revisit rate and early saturation coincide under confinement, the network counterpart of neural text degeneration~\cite{holtzman_degeneration_2020}. Within the Llama family distillation from reasoning traces raises the number of distinct tokens seven-fold on a network that it barely moves. Readings of hallucination as a continuation that leaves the domain sub-graph licensed by the prompt, and of subliminal learning~\cite{cloud_subliminal_2026} and emergent misalignment~\cite{betley_emergentmisalignment_2026} as changes of sequence on an inherited network, remain hypotheses for future investigation.


In summary, an LLM's pretrained embeddings define a fragmented token network onto which generation can be mapped, although the sequence does not follow its edges. The network supplies a cutoff tied to the onset of connectivity, a fingerprint that pre-training fixes and distillation barely moves, a placement of each generation in a concentrated region, and a repetition diagnostic based on how broadly the generation covers the network. These observables are computed from released weights and the generated sequence. A running count of newly visited tokens, for instance, could flag a generation that has become confined to a small sub-graph while it is still being produced. Building a similarity network from learned embeddings and fixing its cutoff near the onset of connectivity is not specific to language, and could be applied to the learned representations now used across the computational sciences~\cite{AI_science_2023} (\blue{Supplementary Note~5}). 
The perspective of token graph paves new way to explain the physical mechanism of LLM generation and demonstrate the domain-specific feature, which motivates the development of cutting-edge domain-adaptive AI models or prompt engineering tools.

The fingerprint, the distillation invariance and the path observables all live on the static similarity graph. Coupling the path to attention deserves future efforts. Lindsey \textit{et al.}~\cite{lindsey_biology_2025} traced planning features that hold several candidate end-words of a poem line at once and steer the line towards them, so generation is locally goal-conditioned over the span about to be emitted. Because consecutive tokens do not follow similarity edges, tools that assume motion along the graph, such as static shortest paths, do not apply directly. Strategic path optimization is required.
Community-aware measures of coverage and of boundary crossing are the natural network tools for the mapped sequences. Beyond the inference-time monitors, the distillation-invariant fingerprint offers a lightweight handle for model auditing, and the measured effective depth informs how many layers a given task warrants. Extending the construction to the context-dependent attention computation, to multilingual and multimodal token spaces, and to a theory of how the fragmented cluster structure forms during pre-training are natural next steps, as is an interventional test in which input embeddings are perturbed along a cluster and the predicted change in revisit rate is measured.

Building a similarity network from learned embeddings, fixing its cutoff near the onset of connectivity and reading generation as coverage of the resulting graph resonates with the network representations now reshaping materials science, where elements, compositions and the literature have been cast as graphs to expose design rules and accelerate the discovery of amorphous alloys~\cite{zhang_constructing_2025,ouyang_graph_2025,atom2vec_pnas_2018}. Characterising the token networks onto which LLM generation is mapped, and steering generation towards under-explored yet promising regions, may extend their role from summarisation to hypothesis generation and materials design.

\clearpage
\section*{Methods}

\subsection*{Models and token embeddings}
We studied six open-source LLMs from four developers (Table~\ref{tab:models}): the pretrained base model Meta-Llama-3.1-8B; Qwen3-8B, post-trained with a hybrid thinking mode; DeepSeek-R1-0528-Qwen3-8B, obtained by distilling DeepSeek-R1-0528 reasoning traces into Qwen3-8B-Base; DeepSeek-R1-Distill-Llama-8B and DeepSeek-R1-Distill-Qwen-7B, distilled from DeepSeek-R1 into Llama-3.1-8B and Qwen2.5-Math-7B; and the reasoning model gpt-oss-20B. For each model we extracted the pretrained input-embedding matrix~\cite{mikolov_word2vec_2013} $\mathbf{E}\in\mathbb{R}^{\mathcal{V}\times d}$, whose row $\mathbf{e}_i\in\mathbb{R}^{d}$ is the embedding of token $i$ in a vocabulary $\mathcal{T}=\{1,\dots,\mathcal{V}\}$. Architectural parameters were read directly from each model's configuration and are reported per model in the Supplementary Information.

\subsection*{Token-similarity network and topology analysis}
For each ordered pair of tokens $(i,j)$ with $i\neq j$ we computed the cosine similarity of their input embeddings,
\begin{equation}
s_{ij} \;=\; \frac{\mathbf{e}_i\!\cdot\!\mathbf{e}_j}{\lVert\mathbf{e}_i\rVert_2\,\lVert\mathbf{e}_j\rVert_2},
\label{eq:cosine}
\end{equation}
and connected the pair when $s_{ij}$ exceeded a threshold $\mathcal{S}\in[0,1]$. This defines an undirected, unweighted graph $G(\mathcal{S})=(\mathcal{T},\mathcal{E}(\mathcal{S}))$ with adjacency
\begin{equation}
A_{ij}(\mathcal{S}) \;=\; \mathbb{1}\!\left[\,s_{ij}>\mathcal{S}\,\right]\quad (i\neq j),\qquad A_{ii}=0,
\label{eq:adjacency}
\end{equation}
edge count $\mathcal{N}_\mathrm{edge}(\mathcal{S})=\tfrac{1}{2}\sum_{i\neq j} A_{ij}(\mathcal{S})$ and node degree $k_i(\mathcal{S})=\sum_{j} A_{ij}(\mathcal{S})$. We swept $\mathcal{S}$ from 0.05 to 0.95 in steps of 0.05 and, at each value, recorded the edge count and the degree of every node. The reference cutoff $\mathcal{S}_\mathrm{c}$ was set where the mean degree
\begin{equation}
\langle k\rangle(\mathcal{S}) \;=\; \frac{1}{\mathcal{N}_\mathrm{node}}\sum_{i=1}^{\mathcal{N}_\mathrm{node}} k_i(\mathcal{S}) \;=\; \frac{2\,\mathcal{N}_\mathrm{edge}(\mathcal{S})}{\mathcal{N}_\mathrm{node}}
\label{eq:meandeg}
\end{equation}
crosses unity, the Erd\H{o}s--R\'enyi percolation criterion, equivalently $\mathcal{N}_\mathrm{edge}=\mathcal{N}_\mathrm{node}/2$ for the unordered-pair edge count used throughout, where $\mathcal{N}_\mathrm{node}=\mathcal{V}$ counts all vocabulary tokens (isolated ones included),
\begin{equation}
\mathcal{S}_\mathrm{c} \;=\; \inf\!\left\{\mathcal{S}\in[0,1]:\;\langle k\rangle(\mathcal{S})\le 1\right\};
\label{eq:percolation}
\end{equation}
$\mathcal{S}_\mathrm{c}$ was located on the sorted edge list (below), which agrees within $0.01$ with linear interpolation of $\langle k\rangle(\mathcal{S})$ between the two swept points bracketing $\langle k\rangle=1$. All sub-graph and fingerprint analyses are performed at a per-model \emph{working cutoff} $\mathcal{S}^{*}$: the sweep grid point at which the released graphs were built, lying $0.015$--$0.052$ below $\mathcal{S}_\mathrm{c}$. This approximation is made for the convenience of investigation. There the mean degree is $\langle k\rangle=1.4$--$2.6$, close to the susceptibility peak, and the LCC holds $2$--$13\%$ of the vocabulary.

Following the onset of connectivity, we extract the LCC or subgraph from the token graph. We measured its size $\mathcal{N}_\mathrm{LCC}$, the fraction of non-isolated nodes and the susceptibility $\chi=\bigl(\sum_{\mathcal{C}}|\mathcal{C}|^2-\mathcal{N}_\mathrm{LCC}^2\bigr)/\mathcal{N}_\mathrm{node}$, the mean size of the clusters (connected components) $\mathcal{C}$ other than the largest, as functions of both $\mathcal{S}$ and $\langle k\rangle$.
For Qwen3-8B the LCC sizes on the sweep grid agree exactly with an independent sparse-matrix computation of the pipeline. Two null embeddings with the same $\mathcal{V}$ and $d$ were analysed identically: rows drawn from a Gaussian with the model's mean and covariance, and the model's embedding matrix with each column permuted independently across tokens. 
This feature indicates the probable training inefficiency as the parameters are not fully optimized and the process is not converged to a local optimal.
Null degree distributions were compared at the edge count of the model at $\mathcal{S}^{*}$. 

Token strings were decoded from each model's byte-level vocabulary. A token was classed as a word token if its decoded string contains a letter, and its surface-form key was the decoded string with surrounding whitespace removed and letters lower-cased. At $\mathcal{S}^{*}$ we recorded the class of the 100 highest-degree tokens, the fraction of edges joining two tokens with the same key and the fraction of finite clusters (connected components of two or more tokens other than the LCC) whose tokens all share one key. Two controls re-evaluated the LCC fraction: the sub-graph induced on word tokens, compared with a Gaussian null fitted to the word-token embeddings and thresholded at the same edge count; and the graph obtained by merging all tokens with the same key into one node, compared with the full-vocabulary Gaussian null at matched edge count, merged by the same key assignment.

\subsection*{Graph statistics}
For each model we computed the number of components, the largest component $G_{\mathrm{LCC}}$ of size $\mathcal{N}_\mathrm{LCC}$, the mean degree (eq.~\ref{eq:meandeg}), the degree distribution
\begin{equation}
P(k) \;=\; \frac{1}{\mathcal{N}_\mathrm{node}}\sum_{i=1}^{\mathcal{N}_\mathrm{node}}\mathbb{1}[\,k_i=k\,],
\label{eq:pk}
\end{equation}
the modularity, and the number of communities. Communities $\{C_1,\dots,C_{n_c}\}$ were detected by greedy modularity maximisation, and the modularity itself was evaluated as
\begin{equation}
Q \;=\; \frac{1}{2\,\mathcal{N}_\mathrm{edge}}\sum_{i,j}\!\left[A_{ij}-\frac{k_i k_j}{2\,\mathcal{N}_\mathrm{edge}}\right]\delta\!\bigl(c_i,c_j\bigr),
\label{eq:modularity}
\end{equation}
where $c_i$ is the community of node $i$ and $\delta(\cdot,\cdot)$ is the Kronecker delta~\cite{newman_structure_2003}. Network fingerprints used three statistics of the largest connected component $G_\mathrm{LCC}$ --- its mean degree $\langle k\rangle_\mathrm{LCC}$ (eq.~\ref{eq:meandeg} restricted to $G_\mathrm{LCC}$), its modularity $Q$ and its community fraction $f_c=n_c/\mathcal{N}_\mathrm{LCC}$ --- $z$-scored across models,
\begin{equation}
\tilde x_m^{(j)} \;=\; \frac{x_m^{(j)}-\mu_j}{\sigma_j},\qquad \mathbf{f}_m=\bigl(\tilde{\langle k\rangle}_{\mathrm{LCC},m},\,\tilde Q_m,\,\tilde f_{c,m}\bigr),
\label{eq:fingerprint}
\end{equation}
with $\mu_j,\sigma_j$ the mean and standard deviation of feature $j\in\{\langle k\rangle_\mathrm{LCC},Q,f_c\}$ over the six models; the distance matrix is the pairwise Euclidean distance
\begin{equation}
D_{mn} \;=\; \lVert\mathbf{f}_m-\mathbf{f}_n\rVert_2,
\label{eq:distance}
\end{equation}
ordered by average-linkage hierarchical clustering. 

\subsection*{Layer-wise generation and the logit lens}
Each model was run on nine prompts, three topics (metallic glasses, machine learning and movies) at three complexities (simple, moderate and complex), listed in Supplementary Table~7. Generations used the token-by-token method with a key--value cache. For every generated position $t\in\{1,\dots,T_\mathrm{gen}\}$ we saved, for each layer $\ell\in\{0,\dots,L-1\}$ (0-based from the first transformer block, so $\ell=L-1$ is the deepest layer), the top-$K$ token identities and probabilities ($K=256$) obtained by projecting the layer's hidden state $\mathbf{h}^{(\ell)}_t\in\mathbb{R}^{d}$ through the shared final normalisation $\mathrm{RMSNorm}(\cdot)$ and output head $W_U\in\mathbb{R}^{\mathcal{V}\times d}$ (the logit lens~\cite{nostalgebraist_logitlens_2020}). 
$p^{(L-1)}_t$ is the model's own output distribution,
\begin{equation}
\mathbf{z}^{(\ell)}_t \;=\; W_U\,\mathrm{RMSNorm}\bigl(\mathbf{h}^{(\ell)}_t\bigr),\qquad
p^{(\ell)}_{t,v} \;=\; \frac{\exp\bigl(z^{(\ell)}_{t,v}\bigr)}{\sum_{u\in\mathcal{T}}\exp\bigl(z^{(\ell)}_{t,u}\bigr)}.
\label{eq:logitlens}
\end{equation}
From these we computed, per position, the layer-wise top-1 prediction $\hat v^{(\ell)}_t=\arg\max_{v} p^{(\ell)}_{t,v}$ and three observables. The settling layer, which corresponds to the saturation event of Lioubashevski \textit{et al.}~\cite{lioubashevski_top1_2025}, is the earliest layer at and beyond which the layer-wise top-1 prediction equals the deepest-layer top-1 and remains stable,
\begin{equation}
\ell^{\star}_t \;=\; \min\!\left\{\ell:\;\hat v^{(\ell')}_t=\hat v^{(L-1)}_t\;\;\forall\,\ell'\ge\ell\right\},
\label{eq:settling}
\end{equation}
whose target coincides, under greedy decoding, with the token the model actually emits. We report it as the settling depth $\ell^{\star}_t/(L-1)\in[0,1]$, write $\bar\ell^{\star}$ for its mean over the generated positions of a run, and say that a token settles in the top quartile of layers when $\ell^{\star}_t/(L-1)\ge0.75$. The per-layer Shannon entropy, evaluated on the renormalised truncated top-$K$ distribution $\tilde p^{(\ell)}_t$, is
\begin{equation}
H^{(\ell)}_t \;=\; -\!\!\sum_{v\in\mathrm{Top}_K^{(\ell)}(t)}\!\!\tilde p^{(\ell)}_{t,v}\,\log\tilde p^{(\ell)}_{t,v},
\label{eq:entropy}
\end{equation}
and the per-layer Kullback--Leibler divergence to the final-layer distribution, evaluated on the union $\mathcal{U}^{(\ell)}_t=\mathrm{Top}_K^{(\ell)}(t)\cup\mathrm{Top}_K^{(L-1)}(t)$ with missing entries floored at $\varepsilon=10^{-9}$ and renormalised into $\hat p^{(\ell)}_t$, is
\begin{equation}
\mathrm{KL}\!\bigl(\hat p^{(\ell)}_t\,\Vert\,\hat p^{(L-1)}_t\bigr) \;=\; \sum_{v\in\mathcal{U}^{(\ell)}_t} \hat p^{(\ell)}_{t,v}\log\frac{\hat p^{(\ell)}_{t,v}}{\hat p^{(L-1)}_{t,v}}.
\label{eq:kl}
\end{equation}
All logarithms are natural, so entropies and divergences are in nats.
Generated sequences were mapped onto the prompt-and-generated token sub-graph by locating each token as a node and adding a directed sequence step between consecutive tokens, $\{(v_t\to v_{t+1})\}_{t=1}^{T_\mathrm{gen}-1}$. These steps are drawn on top of the similarity edges and are not themselves similarity edges unless $A_{v_t v_{t+1}}=1$.

\subsection*{Transition-edge test}
For a generated sequence $\mathbf{v}=(v_1,\dots,v_{T_\mathrm{gen}})$, let $\mathcal{P}=\{t: v_t\neq v_{t+1}\}$ be the set of transitions between distinct tokens. The transition-edge fraction is
\begin{equation}
\phi(\mathbf{v}) \;=\; \frac{1}{|\mathcal{P}|}\sum_{t\in\mathcal{P}} A_{v_t v_{t+1}}(\mathcal{S}^{*}).
\label{eq:transition}
\end{equation}
Its baseline is $\phi$ evaluated on 1,000 random permutations of $\mathbf{v}$, which keep the multiset of tokens of the run and remove its order.
We also computed $\phi$ for pairs drawn independently from the run's unigram distribution, which gives the same values, and the graph density $\langle k\rangle/(\mathcal{V}-1)$. The $95\%$ interval of the observed $\phi$ is a moving-block bootstrap over transitions (block length 50, 2,000 resamples). Runs were pooled per model by concatenating their transitions. The same statistics were computed for five temperature-$0.6$ runs of Qwen3-8B and gpt-oss-20B on the moderate machine-learning prompt, generated with the same pipeline for a companion temperature study.

For Qwen3-8B, whose input embedding and output head are untied, every row $\mathbf{e}_i$ of the input-embedding matrix was replaced by $\lVert\mathbf{e}_i\rVert(c\,\hat{\mathbf{e}}_i+\sqrt{1-c^2}\,\mathbf{u}_i)$ with $c=0.99$ and $\mathbf{u}_i\perp\hat{\mathbf{e}}_i$ a unit vector, so that every token keeps its norm and has cosine $c$ with its original embedding. In the random edit $\mathbf{u}_i$ is an independent Gaussian direction. In the targeted edit it points away from the sum of the unit embeddings of the token's neighbours at $\mathcal{S}^{*}$ (isolated tokens receive the random direction). The output head was not changed, so the logit lens is unaffected. 
All statistics were evaluated on the unedited network.

\subsection*{Domain sub-graphs and path metrics}
For each domain we used a simple prompt (metallic glasses, machine learning, movies) and built the sub-graph induced by the prompt and generated tokens together with their similarity neighbours, augmented by the sequence steps. Communities were detected by greedy modularity maximisation of $Q$ (eq.~\ref{eq:modularity}). Domain overlap is the Jaccard index of the activated token sets $\mathcal{T}^{(a)},\mathcal{T}^{(b)}\subseteq\mathcal{T}$,
\begin{equation}
J\!\bigl(\mathcal{T}^{(a)},\mathcal{T}^{(b)}\bigr) \;=\; \frac{\bigl|\mathcal{T}^{(a)}\cap\mathcal{T}^{(b)}\bigr|}{\bigl|\mathcal{T}^{(a)}\cup\mathcal{T}^{(b)}\bigr|}.
\label{eq:jaccard}
\end{equation}
Given a generated token sequence $\mathbf{v}$, repetition was characterised by the token-revisit rate, the distinct-$n$-gram ratio (with $n=3$), and the cumulative count of unique tokens over the run,
\begin{equation}
r_\mathrm{rev}(\mathbf{v}) \;=\; 1 - \frac{\bigl|\{v_t\}_{t=1}^{T_\mathrm{gen}}\bigr|}{T_\mathrm{gen}},\qquad
\rho_n(\mathbf{v}) \;=\; \frac{\bigl|\{(v_t,\dots,v_{t+n-1})\}_{t=1}^{T_\mathrm{gen}-n+1}\bigr|}{T_\mathrm{gen}-n+1},
\label{eq:revisit_ngram}
\end{equation}
\begin{equation}
U(\tau) \;=\; \bigl|\{v_t\}_{t=1}^{\tau}\bigr|,\qquad \tau\in\{1,\dots,T_\mathrm{gen}\}.
\label{eq:unique}
\end{equation}
Across-model summary statistics (e.g.\ the Pearson correlation in Fig.~\ref{fig:quality}c) were computed on the per-model pairs $\bigl(r_\mathrm{rev}^{(m)},\rho_3^{(m)}\bigr)$.


\subsection*{Data availability}
The models are open-sourced.
An interactive dashboard visualising the model features and token generation on the graph is available at \href{https://ychu-mining-ai-dashboard.hf.space/}{https://ychu-mining-ai-dashboard.hf.space/}.

\subsection*{Code availability}
The code pipelines are available from the authors upon reasonable request.


\subsection*{Author contributions}

Y.-C.H. conceived the project, performed the analyses and wrote the manuscript.

\subsection*{Competing interests}

The authors declare no competing interests.

\section*{Acknowledgements}

We thank the developers of the open-weight models analysed here for releasing their weights. 
Generative AI assisted with literature organization, manuscript editing, and the development and review of plotting and figure-generation scripts.
The author directed its use and independently verified all scientific claims, calculations, code, figures, and references.
No competing interests exist.


\end{spacing}
\end{document}